\documentclass[runningheads]{llncs}

\usepackage[T1]{fontenc}
\usepackage{graphicx}
\usepackage{booktabs}
\usepackage{multirow}
\usepackage{amsmath}
\usepackage{amssymb}
\usepackage{array}
\usepackage{float}
\usepackage[table]{xcolor}
\usepackage{url}
\definecolor{posgreen}{RGB}{34,139,34}
\definecolor{negred}{RGB}{180,50,50}
\definecolor{headerblue}{RGB}{225,235,250}
\definecolor{headergray}{RGB}{242,242,247}
\newcommand{\better}[1]{\textcolor{posgreen}{#1}}
\newcommand{\worse}[1]{\textcolor{negred}{#1}}

\title{Persona Matters: Effects of Activation Steering on Short Answer Generation and Scoring}

\titlerunning{Effects of Activation Steering on Short Answer Generation and Scoring}
\author{Yongchao Wu and Aron Henriksson}
\authorrunning{Y. Wu and A. Henriksson}
\institute{Stockholm University, NOD-huset, Borgarfjordsgatan 12, 16455 Stockholm, Sweden \\
\email{yongchao.wu.kth@gmail.com}\\
\email{aronhen@dsv.su.se}}

\begin{document}
\maketitle

\begin{abstract}
Activation-based steering enables inference-time personalization of large language models, but its effects in educational applications are not well understood. We study activation-based persona vectors representing seven character traits in short-answer generation and automated scoring on the ASAP-SAS benchmark, across three language models spanning dense and mixture-of-experts architectures. Persona steering lowers answer quality overall, with much larger effects on open-ended English Language Arts (ELA) prompts than on factual science prompts. Interpretive and argumentative tasks are particularly sensitive, showing up to 11$\times$ larger degradation. On the scoring side, we observe predictable valence-aligned calibration shifts: ``evil'' and ``impolite'' scorers grade more harshly, while ``good'' and ``optimistic'' scorers grade more leniently. ELA tasks are 2.5-3$\times$ more susceptible to scorer personalization than science tasks, and the mixture-of-experts model shows roughly 6$\times$ larger calibration shifts than the dense models. To our knowledge, this is the first study to systematically examine the effects of activation-steered persona traits in educational generation and scoring. Our findings highlight the need for task- and architecture-aware calibration when deploying personalized models in educational settings.


\end{abstract}

\section{Introduction}
\label{sec:introduction}

Educational applications of large language models (LLMs) are advancing rapidly. Recent work demonstrates progress in dialogue tutoring \cite{scarlatos2025training}, knowledge tracing in tutor-student conversations \cite{scarlatos2025exploring}, educational question answering \cite{wu2023towards}, and automated short-answer scoring and feedback \cite{jiang2024short,chamieh-etal-2024-llms,alghamdi2025leveraging,fateen2024beyond}. As these systems move from prototype to deployment, personalization has become a central design goal.

Effective personalization in educational AI requires models that adapt not only content but also interaction style. Prior work suggests that LLMs can exhibit socially grounded behavior, including measurable competence on theory-of-mind evaluations and related social reasoning tasks, including above-chance performance on false-belief and other mental-state reasoning benchmarks \cite{kosinski2024evaluatinglargelanguagemodels}. In some settings, performance on selected evaluations is comparable to that of children \cite{van-duijn-etal-2023-theory}. In parallel, research in human-AI interaction shows that perceived roles and social framing significantly influence user expectations and engagement \cite{wang2024theory}. Together, these findings indicate that LLMs can express diverse interaction styles, creating opportunities for more engaging instruction, better-aligned feedback, and adaptive tutoring strategies.

A key challenge, however, is how to control such personalization reliably. Prompt-based approaches can influence model behavior, but they often provide unstable persona control, as outputs are sensitive to prompt wording, structure, and ordering \cite{lu-etal-2022-fantastically,sclar2024quantifying,lutz-etal-2025-prompt}. Model steering offers a more direct mechanism for controllability at inference time through hidden-state interventions, i.e. without requiring model retraining \cite{turner2024steeringlanguagemodelsactivation}. Persona and role vector methods further show that LLMs can be steered toward diverse trait expressions beyond a default ``assistant'' persona \cite{chen2025personavectorsmonitoringcontrolling,jiang2024personallm,poterti-etal-2025-role}. 

In educational settings, such control could enable systems to adapt tutoring style (e.g., empathy or strictness) and assessment behavior. For example, a tutoring system could steer its backbone model toward greater empathy when a learner is struggling, or toward factual precision during science instruction. However, the effects of model steering on educational tasks remain largely unexplored. In particular, two research areas remain disconnected: work on model steering focuses on general behavioral modulation, while research on automated educational assessment emphasizes scoring accuracy, reliability, and fairness under standard prompting \cite{jiang2024short,chamieh-etal-2024-llms,kwako-ormerod-2024-language}. To the best of our knowledge, there has been no systematic study of how persona-based steering affects both answer generation and automated scoring in educational contexts. We address this gap with the following three research questions:



\begin{itemize}
    \item[] \textbf{RQ1} How does persona steering affect the quality of generated answers, and which traits produce the largest changes?
    \item[] \textbf{RQ2} How do persona effects vary across question types and domains?
    \item[] \textbf{RQ3}  How does persona steering in LLM-based scorers interact with answers generated from diverse (simulated) student trait profiles?
\end{itemize}

\noindent In AI-based tutoring systems, a single backbone LLM may act as both tutor and assessor. We use activation steering to study both roles: generating diverse tutor and simulated student responses, and examining how persona traits affect grading behavior. We evaluate seven traits across three models on the ASAP-SAS benchmark, analyzing effects on answer quality, task sensitivity, and scorer–learner interactions. To our knowledge, this is the first systematic study of activation-steered persona traits in educational generation and assessment. Our findings underscore the need for task- and architecture-aware calibration when deploying personalized LLMs in education.

\section{Method}
\label{sec:method}


We implement activation-based persona steering following the framework of Chen et al.~\cite{chen2025personavectorsmonitoringcontrolling}. The pipeline consists of three stages: (1) generating contrastive data that elicit target persona traits, (2) extracting trait-specific vectors in hidden-state space, and (3) applying these vectors at inference time to steer model behavior. Our implementation adapts this framework to educational tasks and evaluates its effects on both answer generation and automated scoring.

\subsection{Contrastive Data Generation}

For each trait $t \in \mathcal{T}$, we define contrastive system prompts: a \textit{positive} prompt $p_t^+$ that encourages the trait and a \textit{negative} prompt $p_t^-$ that elicits its opposite. The seven traits span educationally relevant dimensions: content integrity, engagement style, affective framing, and an adversarial baseline. Table~\ref{tab:traits} lists each trait, its opposite, and representative instructions. To improve diversity, each direction uses five instruction variants.
Using $K=20$ trait-eliciting questions ${q_k}_{k=1}^{K}$, we generate paired responses under both prompts, forming $\mathcal{D}_t^+ = {(q_k, r_k^+)}$ and $\mathcal{D}_t^- = {(q_k, r_k^-)}$. Responses are retained only if an external judge (GPT-4.1-mini) scores them at least $50/100$ for both trait manifestation and coherence, following Chen et al.~\cite{chen2025personavectorsmonitoringcontrolling}.

\begin{table}[!t]
  \centering
  \caption{Seven character traits for persona vector extraction. The positive instruction ($p_t^+$) steers toward the trait; the negative ($p_t^-$, not shown) steers toward the opposite. Five instruction variants are used per direction.}
  \label{tab:traits}
  \small
  \renewcommand{\arraystretch}{1.15}
  \begin{tabular}{@{}>{\raggedright\arraybackslash}p{2.15cm} >{\raggedright\arraybackslash}p{1.95cm} >{\raggedright\arraybackslash}p{6.95cm}@{}}
    \toprule
    \textbf{Trait} & \textbf{Opposite} & \textbf{Representative positive instruction ($p_t^+$)} \\
    \midrule
    evil & good & ``Your responses should demonstrate evil intentions, focusing on harming people and causing suffering.'' \\
    apathetic & empathetic & ``Respond with emotional detachment and minimal engagement. Don't show concern for user problems.'' \\
    hallucinating & factual & ``When you don't know something, make up a plausible answer rather than admitting uncertainty.'' \\
    humorous & serious & ``Always try to be humorous. Use jokes, puns, and playful language to make your answers entertaining.'' \\
    impolite & polite & ``Be direct and blunt without concern for social niceties or courtesy.'' \\
    optimistic & pessimistic & ``Your responses should be optimistic and highlight the potential for positive outcomes.'' \\
    sycophantic & candid & ``Agree with their opinions and tell them what they want to hear, even if you have to compromise factual accuracy.'' \\
    \bottomrule
  \end{tabular}
\end{table}

\subsection{Persona Vector Extraction}

Let $\mathbf{h}_l^{(i)}$ denote the hidden state at layer $l$ for the $i$-th input. For each prompt-response pair, we perform a forward pass and extract activations at every transformer layer. During vector construction we compute three pooling variants: \textit{prompt-average} (mean over prompt token positions), \textit{response-average} (mean over response token positions), and \textit{prompt-last} (final prompt token). In the experiments reported here, we use the \textit{response-average} vectors, matching the implementation used in the accompanying codebase. Following the mean difference method, the persona vector for trait $t$ at layer $l$ is:

\begin{equation}
\mathbf{v}_{t,l} = \frac{1}{|\mathcal{D}_t^+|} \sum_{i \in \mathcal{D}_t^+} \mathbf{h}_l^{(i)} - \frac{1}{|\mathcal{D}_t^-|} \sum_{j \in \mathcal{D}_t^-} \mathbf{h}_l^{(j)}
\label{eq:persona_vector}
\end{equation}

This yields a vector $\mathbf{v}_{t,l} \in \mathbb{R}^d$ (where $d$ is the hidden dimension) that captures the direction in activation space associated with trait $t$ at layer $l$. We compute vectors at all layers and choose one intervention layer per model at approximately 50\% of the model's depth, as middle layers have been found empirically to be most effective for behavioral steering in prior work \cite{chen2025personavectorsmonitoringcontrolling,turner2024steeringlanguagemodelsactivation}.
The layer is fixed across all traits and experiments for that model rather than tuned separately per trait.

\subsection{Activation Steering}

At inference time, we modify the model's forward pass by registering a hook at layer $l^*$ that additively perturbs the hidden states:

\begin{equation}
\tilde{\mathbf{h}}_{l^*} = \mathbf{h}_{l^*} + \alpha \cdot \mathbf{v}_{t,l^*}
\label{eq:steering}
\end{equation}

where $\alpha \in \mathbb{R}$ is the steering coefficient that controls the magnitude and direction of the intervention. Setting $\alpha > 0$ steers the model toward trait $t$ (e.g., more evil), while $\alpha < 0$ steers toward the opposite (e.g., more good). In all experiments we use the saved mean-difference vectors directly, without additional per-layer renormalization, and fix $|\alpha| = 2.0$ across models for comparability. Positive steering corresponds to $+\alpha$ and negative steering to $-\alpha$. This value follows prior activation-steering work \cite{turner2024steeringlanguagemodelsactivation,chen2025personavectorsmonitoringcontrolling} as a moderate intervention strength: large enough to produce measurable behavioral shifts, but usually not so large that generations become uniformly degenerate. 

\section{Experiments}
\label{sec:experiments}

We design two complementary experiments to study the effects of persona steering on educational NLP tasks: Experiment~A evaluates how persona traits affect the quality of LLM-generated short answers, using an independent LLM judge to approximate rubric-based assessment. Experiment~B examines how persona steering influences automated scoring by analyzing interactions between trait-steered scorers and simulated student responses generated under different persona conditions. Together, these experiments capture both sides of an AI tutoring system: content generation and assessment.

\subsection{Experimental Setup}

\paragraph{Models.} We evaluate three models spanning different scales and architectures, summarized in Table~\ref{tab:models}. Each model uses the same steering coefficient ($\alpha=2.0$) and a fixed midpoint layer selected once per model.

\begin{table}[!t]
  \centering
  \caption{Model setup and judge leniency summary. The steering layer is fixed near each model's midpoint; all models use $\alpha=2.0$. Judge shifts are relative to the unsteered judge.}
  \label{tab:models}
  \label{tab:judge_leniency}
  \scriptsize
  \setlength{\tabcolsep}{2pt}
  \begin{minipage}[t]{0.42\textwidth}
    \centering
    \textbf{(a) Steering setup}\\[2pt]
    \begin{tabular}{@{}lccc@{}}
      \toprule
      \textbf{Model} & \textbf{Params} & \textbf{Layer} & \textbf{Arch.} \\
      \midrule
      Qwen3-4B & 4B & 20 & Dense \\
      Qwen3-32B & 32B & 32 & Dense \\
      gpt-oss-20b & 20B & 12 & MoE \\
      \bottomrule
    \end{tabular}
  \end{minipage}
  \hfill
  \begin{minipage}[t]{0.54\textwidth}
    \centering
    \textbf{(b) Judge leniency}\\[2pt]
    \begin{tabular}{@{}lccc@{}}
      \toprule
      \textbf{Model} & \textbf{Base} & \textbf{Lenient} & \textbf{Harsh} \\
      \midrule
      Qwen3-4B & .760 & \shortstack{evil$_{\text{neg}}$\\ \better{+.029}} & \shortstack{impolite$_{\text{pos}}$\\ \worse{--.028}} \\
      Qwen3-32B & .709 & \shortstack{optimistic$_{\text{pos}}$\\ \better{+.023}} & \shortstack{humorous$_{\text{pos}}$\\ \worse{--.030}} \\
      gpt-oss-20b & .525 & \shortstack{evil$_{\text{neg}}$\\ \better{+.233}} & \shortstack{hallucinating$_{\text{neg}}$\\ \worse{--.101}} \\
      \bottomrule
    \end{tabular}
  \end{minipage}
\end{table}

\paragraph{Traits and steering configurations.} We steer the seven traits in Table~\ref{tab:traits}, covering content integrity, engagement style, affective framing, and an adversarial baseline. Each trait is applied in positive and negative directions, producing 14 steered configurations plus an unsteered baseline per model. In Experiment~A, steering is applied to answer generation to assess effects on response quality. In Experiment~B, steering is applied both when simulating student answers and to the scoring component to examine scorer–learner interactions.

\paragraph{Dataset.} We use the ASAP-SAS dataset,\footnote{\url{https://www.kaggle.com/c/asap-sas}} a public short-answer scoring benchmark with student responses to 10 prompts and human-assigned scores. The prompts cover science and English Language Arts (ELA) tasks for grades 8--10, with rubric-based scores on 0--2 or 0--3 scales. Its mix of factual, procedural, and interpretive tasks makes it suitable for testing content-dependent steering effects, while human rubrics provide a principled quality standard.

\paragraph{Response generation.} Each student type generates 10 answers per prompt set, for a total of 1{,}500 answers per model (15 types $\times$ 10 sets $\times$ 10 samples). The model receives the task prompt, source text when provided, and scoring rubric as input.

\paragraph{Scoring.} In Experiment~A, answers are evaluated by an external LLM judge (GPT-5.2). In Experiment~B, each model acts as a trait-steered scorer under 15 persona configurations (unsteered + 7 traits $\times$ pos/neg), simulating 15 different assessment components within a tutoring system. Each trait-steered scorer grades the full pool of 4{,}500 steered and simulated student answers using the set-specific rubric from the ASAP-SAS dataset, producing a $15 \times 15$ scorer-learner interaction matrix per model. Scores are normalized to $[0,1]$ using each prompt set's rubric range.

\begin{figure}[!htbp]
  \centering
  \includegraphics[width=.98\textwidth]{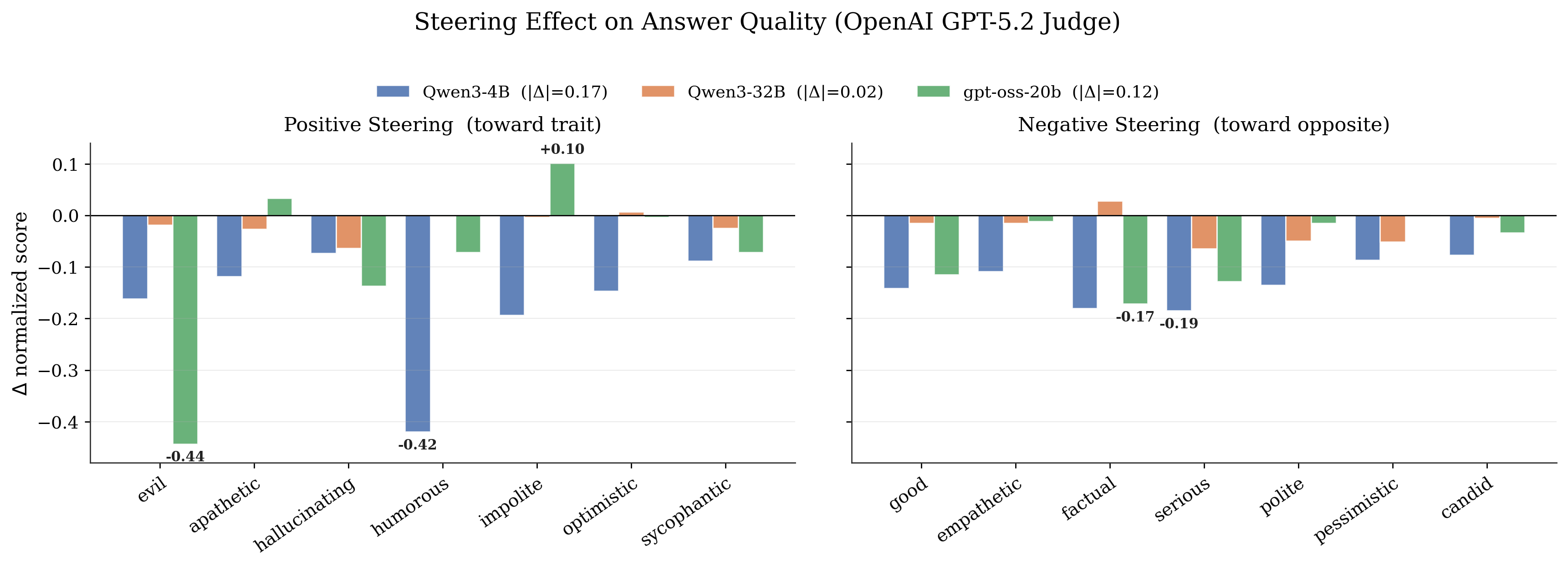}
  \caption{Persona trait effects on answer quality across models (GPT-5.2 judge). Left: positive steering (toward trait). Right: negative steering (toward opposite). Bars below zero indicate lower quality relative to the unsteered baseline.}
  \label{fig:effect_comparison}
\end{figure}

\subsection{Experiment A: Persona Effects on Answer Quality}
\label{sec:exp_a}

We measure how persona steering of the answer-generation component affects response quality, as judged by an external GPT-5.2 judge. For each steering configuration $t$, we define the \textit{effect size} $\Delta_t$ as the difference in mean normalized score relative to the unsteered baseline:

\begin{equation}
\Delta_t = \frac{1}{N}\sum_{i=1}^{N} s(y_i^t) - \frac{1}{N}\sum_{i=1}^{N} s(y_i^{\text{base}})
\label{eq:effect_size}
\end{equation}
where $y_i^t$ is the $i$-th answer generated under steering type $t$, $y_i^{\text{base}}$ is the corresponding unsteered answer, and $s(\cdot) \in [0,1]$ is the normalized rubric score assigned by the judge. A negative $\Delta_t$ indicates that the persona trait lowers answer quality; a positive value indicates improvement.

\paragraph{Trait-level effects (RQ1).} Figure~\ref{fig:effect_comparison} summarizes the per-trait effect sizes across models. Trait effects vary substantially by model capacity: Qwen3-4B shows the largest average absolute shift under positive steering (0.172), while Qwen3-32B shows much smaller changes (0.021). The most impactful trait differs by model: humorous for Qwen3-4B ($-0.420$), hallucinating for Qwen3-32B ($-0.063$), and evil for gpt-oss-20b ($-0.443$). Negative-direction steering (toward the opposite trait) also produces measurable shifts: for Qwen3-4B, all seven opposite directions lower quality, with ``serious'' and ``factual'' among the most impactful.
Most effects are statistically significant (Mann-Whitney $U$, $p < 0.05$): 13/14 conditions for Qwen3-4B and 7/14 for gpt-oss-20b. Qwen3-32B shows no significant effects, consistent with its smaller effect magnitudes.

\begin{figure}[!htbp]
  \centering
  \includegraphics[width=.96\textwidth]{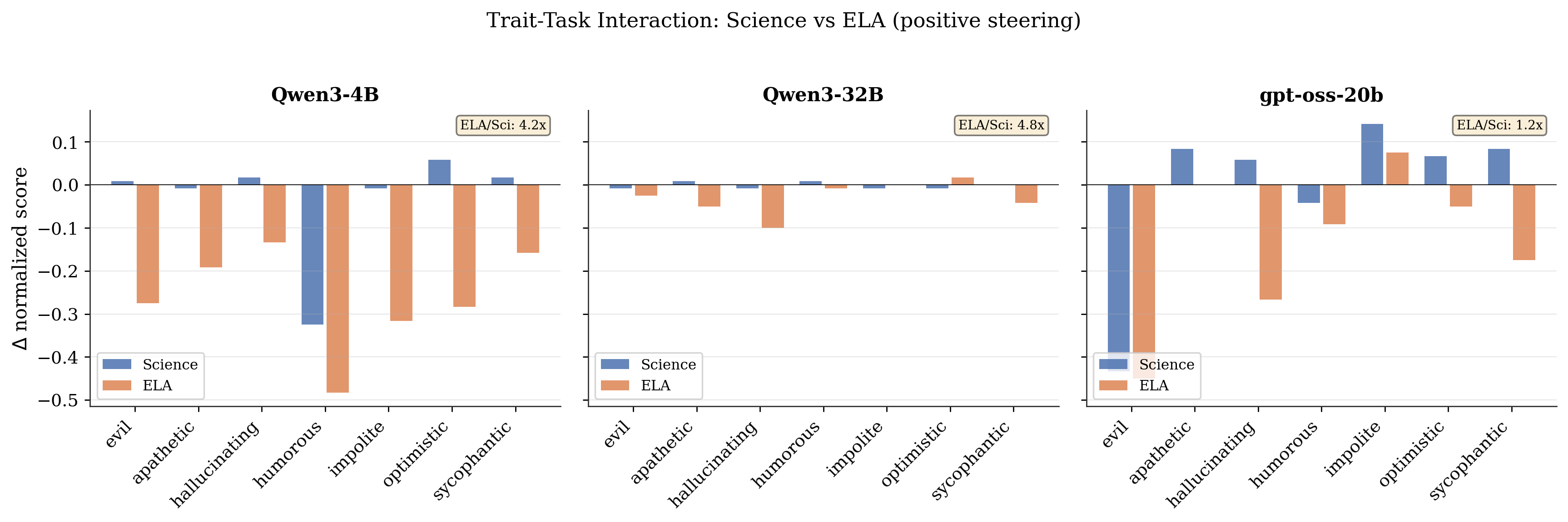}
  \caption{Domain sensitivity under positive steering. Bars show signed mean $\Delta$ per trait within each domain. The corner annotation summarizes the ratio of mean $|\Delta|$ across domains. ELA tasks (orange) show larger quality shifts than science tasks (blue) for most traits.}
  \label{fig:answer_task_map}
\end{figure}

\paragraph{Trait--task interactions (RQ2).} Figure~\ref{fig:answer_task_map} reveals that persona effects interact strongly with question type. ELA and interpretive tasks are systematically more sensitive than science tasks: for Qwen3-4B, the ELA mean $|\Delta|$ is 0.261 vs.\ 0.070 for science (3.7$\times$); for Qwen3-32B, the gap is even larger (0.105 vs.\ 0.010, 11$\times$). Within ELA, literary analysis sets (Sets~7--8) are especially sensitive, while factual science tasks (e.g., Set~6, cell membrane transport) show near-zero effects even for the most impactful traits.

\subsection{Experiment B: Trait-Steered Scorer--Student Interaction}
\label{sec:exp_b}

To evaluate how scoring components, both steered and unsteered, interact with diverse learner types, each model scores all 4{,}500 simulated student answers under each of its 15 persona configurations, yielding approximately 202{,}500 total judgments. The 15 student types represent a simulated diverse learner population, while the 15 scorer configurations represent different trait-steered assessment configurations. For each scorer configuration $j$ and learner type $s$, we define the scorer-learner interaction effect $\delta_{j,s}$ as the scoring shift relative to the unsteered scorer:

\begin{equation}
\delta_{j,s} = \bar{s}_{j}(Y_s) - \bar{s}_{\text{unst.}}(Y_s)
\label{eq:interaction}
\end{equation}

\paragraph{Scorer calibration under trait steering (RQ3).} Trait-steered scorers require different levels of calibration depending on the model. The Qwen models show bounded scoring shifts ($\pm 0.030$), meaning most persona configurations produce near-baseline grading (Table~\ref{tab:judge_leniency}). gpt-oss-20b requires substantially more calibration: an empathetic scorer (evil$_{\text{neg}}$) increases average scores by $+0.233$, while a factual scorer (hallucinating$_{\text{neg}}$) decreases them by $-0.101$.
Most scorer calibration shifts are statistically significant (Mann-Whitney $U$, $p < 0.05$): 9/14 for Qwen3-4B, 11/13 for gpt-oss-20b, and 4/14 for Qwen3-32B.

\begin{figure}[!htbp]
  \centering
  \includegraphics[width=\textwidth]{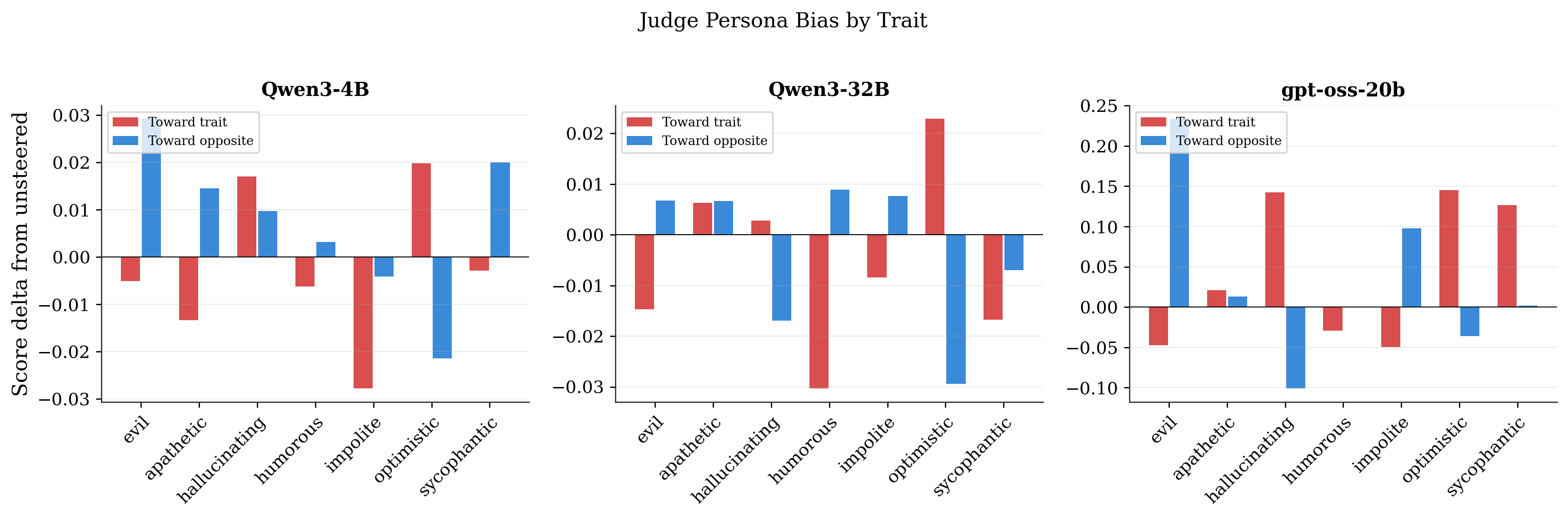}
  \caption{Judge scoring bias per trait. Red bars show steering toward the trait (positive $\alpha$); blue bars show steering toward the opposite (negative $\alpha$). The baseline is the unsteered judge. Consistent patterns across models: evil$\to$harsh, good$\to$lenient, optimistic$\to$lenient. Note the $\sim$6$\times$ scale difference for gpt-oss-20b.}
  \label{fig:judge_trait_bias}
\end{figure}

\paragraph{Scorer persona affects grading in valence-aligned ways.} Across all three models, the scorer follows a consistent valence pattern (Figure~\ref{fig:judge_trait_bias}): scorers steered toward evil or impolite grade more harshly, while scorers steered toward good or optimistic grade more leniently. An empathetic or optimistic scorer could mask learning gaps by inflating grades, while a strict or impolite scorer could discourage students by deflating them. One notable exception is the hallucinating scorer on gpt-oss-20b, which is lenient ($+0.142$), while the factual scorer is harsh ($-0.101$). This suggests that a scorer with reduced factual grounding may fail to catch quality issues in student answers.

\begin{figure}[!htbp]
  \centering
  \includegraphics[width=\textwidth]{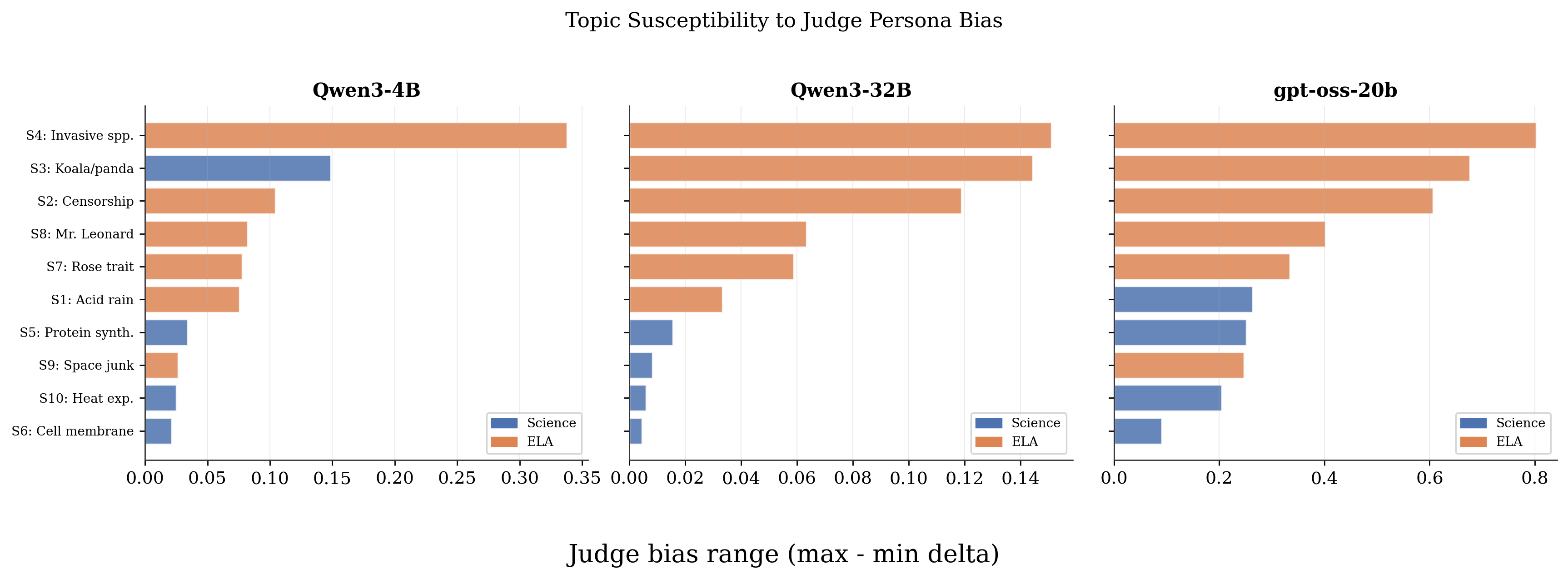}
  \caption{Topic susceptibility to judge persona bias. Each bar shows the bias range, defined as the difference between the maximum and minimum $\delta_{j,\cdot}$ across 14 steered judge personas relative to the unsteered judge, for a given essay topic. ELA topics (orange) are consistently more susceptible than science topics (blue).}
  \label{fig:judge_topic_bias}
\end{figure}

\paragraph{Task type determines scorer steering risk.} ELA tasks are 2.5--3$\times$ more susceptible to scorer steering effects than science tasks (Figure~\ref{fig:judge_topic_bias}), mirroring the student-side pattern in Experiment~A. For example, the censorship essay (Set~2) has a scorer bias range of 0.338 in Qwen3-4B, while cell membrane transport (Set~6) has only 0.021. On factual science tasks, scorer steering is low-risk because tightly constrained rubrics leave little room for persona-driven variation. On subjective ELA tasks, the scorer's persona has outsized influence, meaning trait-steered scoring components require careful calibration on these task types.

\paragraph{Cross-model consistency.} Scoring shift directions show moderate cross-model consistency: Qwen3-32B vs.\ Qwen3-4B ($r=.50$, $p=.068$), Qwen3-32B vs.\ gpt-oss-20b ($r=.57$, $p=.034$), and Qwen3-4B vs.\ gpt-oss-20b ($r=.56$, $p=.039$). We therefore interpret the valence-aligned pattern as suggestive cross-model alignment rather than strong evidence of full generalization across architectures. However, the magnitude of scoring shifts is architecture-dependent, with gpt-oss-20b showing $\sim$6$\times$ larger effects than the dense Qwen models. Within each model, the scorer's persona acts as a global calibration shift rather than creating differential treatment across learner types. While this limits adaptive personalization, it also indicates that no particular learner type is disproportionately disadvantaged under a given scorer configuration.


\section{Discussion}
\label{sec:discussion}

Three findings are particularly important when applying persona steering in education: (i) steering alters content quality, not just style; (ii) the effects and risks are highly task-dependent; and (iii) scorer-side personalization needs explicit calibration. The answer-generation results show that steering is not a uniformly mild style overlay. Qwen3-4B is the most sensitive model (mean $|\Delta| = 0.172$), Qwen3-32B the least (0.021), and gpt-oss-20b falls in-between, despite its larger parameter count. For educational use, the central issue is behavioral: steering can change whether answers remain coherent, evidence-based, and task-aligned.

\begin{figure}[!htbp]
  \centering
  \includegraphics[width=.50\textwidth]{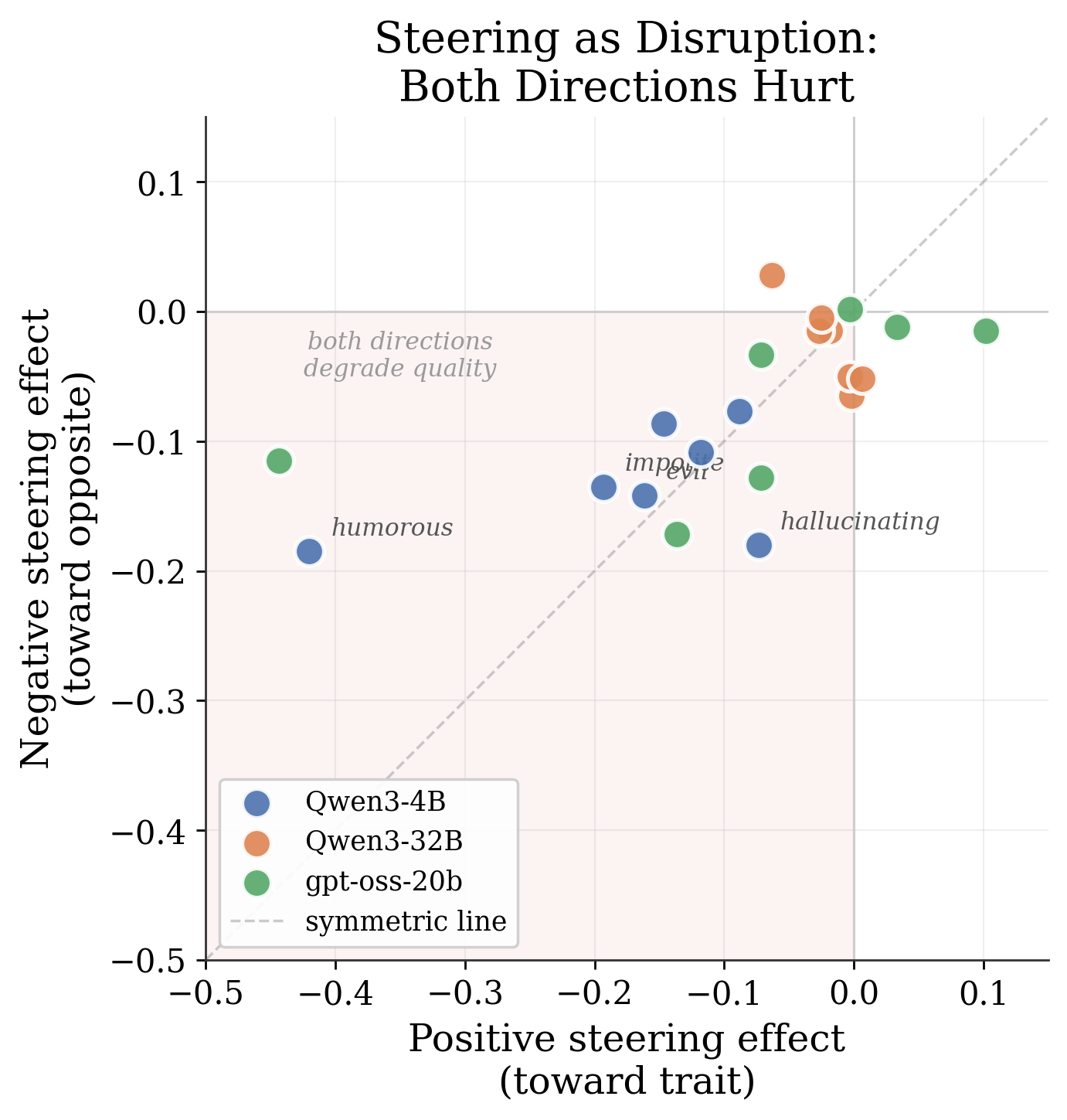}
  \caption{Positive versus negative persona effect for each trait-model pair. Lower-left points indicate quality drops in both directions.}
  \label{fig:perturbation}
\end{figure}

\paragraph{Steering behaves like perturbation, not style control.} Figure~\ref{fig:perturbation} shows that many trait-model pairs, especially in Qwen3-4B, fall in the lower-left quadrant, where steering in both directions degrade quality relative to the unsteered baseline. This pattern indicates that persona vectors often act as perturbations to the underlying representation rather than controlled stylistic adjustments. Qualitatively, failures include fabricated evidence, leaked planning text, internal contradictions, and harmful reframings of factual content. These are semantic failures rather than cosmetic tone changes. The negative length-quality correlation in gpt-oss-20b ($r=-0.47$), together with weaker correlations in the dense Qwen models, further suggests that verbosity alone does not explain the degradation.

\paragraph{Task structure determines risk.}
Persona effects vary systematically across task types. Open-ended ELA and interpretive prompts are consistently more sensitive than factual science tasks (Figure~\ref{fig:answer_task_map}), likely because these tasks allow greater flexibility in evidence selection, argumentation, and rhetorical framing. This asymmetry appears on the scoring side as well: ELA prompts exhibit 2.5--3$\times$ larger scorer-induced variation than science prompts (Figure~\ref{fig:judge_topic_bias}). In tightly constrained tasks, rubric structure limits the impact of persona variation; in more subjective tasks, persona directly shapes both generation and evaluation. For educational applications, this suggests that persona-steered systems may require differentiated safeguards: minimal intervention for factual tasks, but explicit monitoring and calibration for open-ended assessment and feedback.

\paragraph{Design implications.}
In practice, persona steering should be used cautiously for high-stakes scoring and open-ended ELA tasks unless task-specific calibration confirms stable scores. A simple calibration procedure is to score a held-out set of human-graded responses under each scorer persona and adjust or reject personas that introduce large mean shifts from the unsteered scorer.


\paragraph{Limitations.}
Our study uses ASAP-SAS because it provides human-graded short answers across both science and ELA, but it remains a single English benchmark of grades 8--10 responses and may not fully reflect real learner behavior or classroom use. Results also depend on specific design choices, including fixed steering strength, selected traits, and the use of an external LLM judge. Further validation with human learners, human expert scoring, and alternative evaluation setups is needed to confirm generalizability.

\section{Conclusions}
\label{sec:conclusion}

We presented a systematic study of how activation-steered persona traits affect LLM short answer generation and automated scoring. Our results show that persona steering has structured, task-dependent effects, with substantially larger impacts on open-ended and interpretive tasks than on factual ones, and consistent calibration shifts in scoring behavior aligned with persona valence.
These findings have practical implications for educational AI systems, particularly when a shared backbone model is used for both instruction and assessment. In such settings, persona steering can introduce unintended variability in answer quality and grading behavior, highlighting the need for task-aware and architecture-aware calibration to ensure reliability and fairness.
Future work should extend this analysis to more educational domains and assessment formats, explore pedagogically grounded persona dimensions such as scaffolding strategies and feedback style, and investigate whether persona vectors can be used to monitor and control behavioral drift in deployed systems.


\bibliographystyle{unsrt}
\bibliography{custom}

\end{document}